\documentclass[runningheads]{llncs}

\usepackage{eccv}

\usepackage{eccvabbrv}

\usepackage{graphicx}
\usepackage{booktabs}

\usepackage[accsupp]{axessibility}  

\usepackage{hyperref}

\usepackage{orcidlink}
\usepackage{enumitem} 

\usepackage[most]{tcolorbox} 

\newtcolorbox{takeawaybox}{
    colback=gray!10!white,
    colframe=white,
    coltitle=black,
    rounded corners, 
    fonttitle=\bfseries,
}
\begin{document}

\title{HyTAS: A Hyperspectral Image Transformer Architecture Search Benchmark and Analysis} 

\titlerunning{HyTAS: Hyperspectral Transformer Architecture Search}

\author{Fangqin Zhou\inst{1}\orcidlink{0000-0001-9005-232X} \and
Mert Kilickaya\inst{1}\orcidlink{0000-0002-3474-082X} \and
Joaquin Vanschoren\inst{1}\orcidlink{0000-0001-7044-9805} \and
Ran Piao\inst{1}}

\authorrunning{F.Zhou et al.}

\institute{Automated Machine Learning Group, Eindhoven University of Technology, the Netherlands \\
\email{\{f.zhou, j.vanschoren\}@tue.nl}\\
\url{https://github.com/openml-labs}
}

\maketitle

\begin{abstract}
Hyperspectral Imaging (HSI) plays an increasingly critical role in precise vision tasks within remote sensing, capturing a wide spectrum of visual data. Transformer architectures have significantly enhanced HSI task performance, while advancements in Transformer Architecture Search (TAS) have improved model discovery. To harness these advancements for HSI classification, we make the following contributions: \textit{i)} We propose HyTAS, the first benchmark on transformer architecture search for Hyperspectral imaging, \textit{ii)} We comprehensively evaluate $12$ different methods to identify the optimal transformer over $5$ different datasets, \textit{iii)} We perform an extensive factor analysis on the Hyperspectral transformer search performance, greatly motivating future research in this direction. All benchmark materials are available at \href{https://github.com/zhoufangqin/HyTAS/tree/main}{HyTAS}.
  
  \keywords{Hyperspectral Image Classification \and Transformer Architecture Search \and Zero-Cost Proxies \and Benchmark and Analysis}
\end{abstract}

\section{Introduction}
\label{sec:intro}

\vspace{-4mm}

Hyperspectral Imaging (HSI) captures a broad spectrum of electromagnetic wavelengths, providing detailed spectral information beyond conventional RGB images. This rich spectral data has significantly impacted various industries, including agriculture for plant monitoring \cite{dale2013hyperspectral, xie2015detection, furbank2021wheat, wang2019early, nagasubramanian2019plant}, remote sensing for Earth analysis \cite{goetz2009three, applicationremotesensing}, and robotics for enhanced navigation and vision \cite{jakubczyk2022hyperspectral, trierscheid2008hyperspectral}.

\begin{figure}[t]
    \centering
    \includegraphics[width=0.85\textwidth]{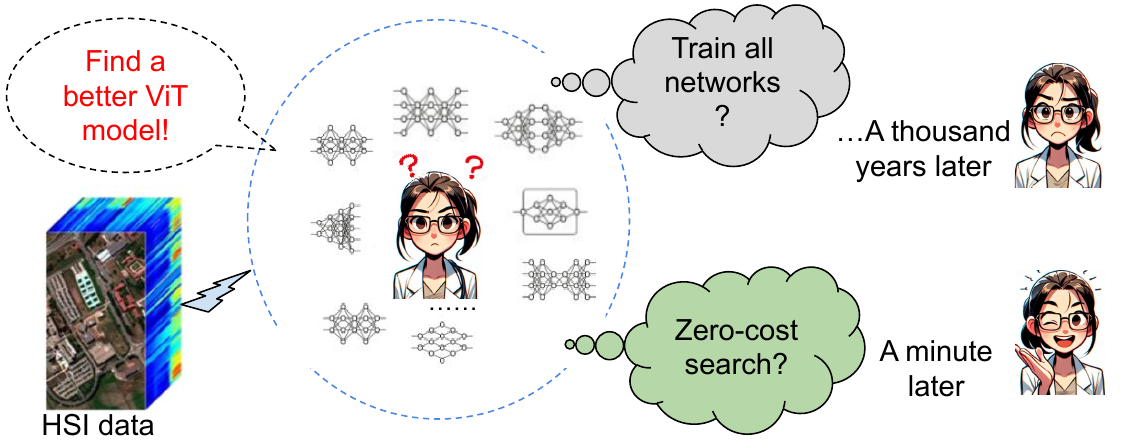}
\caption{A scientist outside AI domain may find it difficult to design a novel Hyperspectral image transformer tailored to their tasks and data. To automate the design of Hyperspectral transformers, we introduce \textbf{Hy}perspectral \textbf{T}ransformer \textbf{A}rchitecture \textbf{S}earch (HyTAS) in this paper. The scientist images are generated by DALL-E.}
\label{fig:motivation}
\end{figure}

Scientists in fields like agriculture or remote sensing may seek to analyze Hyperspectral images using transformer models, see Fig.~\ref{fig:motivation}. However, designing high-performing Hyperspectral image transformers requires significant expertise and computational resources. Transformer Architecture Search (TAS) methods like AutoFormer \cite{chen2021autoformer} or ViTAS \cite{su2022vitas} can help but demand extensive computational resources (e.g., 24 GPU days for AutoFormer).

An alternative is using training-free architecture search techniques, known as zero-cost proxies \cite{snip, synflow, grasp, dss, li2023zico}. Hereafter, we will use the term "proxy" for brevity. Proxies quickly evaluate transformer architectures without training, offering two key advantages: \textit{i) Efficiency:} Proxies identify high-performing transformers within minutes. \textit{ii) Data Independence:} Many proxies do not require real data, reducing setup and data collection costs \cite{synflow, logsynflow, naswot, dss}. This characteristic is especially advantageous for hyperspectral imaging applications, where the setup and calibration of the camera, as well as the collection and annotation of data, incur significant costs.

To that end, in this paper, we introduce \textbf{Hy}perspectral \textbf{T}ransformer \textbf{A}rchitecture \textbf{S}earch (HyTAS), to automatically identify Hyperspectral transformers tailored for downstream tasks and datasets. Our motivations are three-fold: \textit{i) \textbf{Democratization}}: HyTAS enables researchers, especially in fields like biology or agriculture, to identify top-performing, lightweight models using minimal resources. \textit{ii) \textbf{Attention}}: We aim to highlight the unique challenges of HSI to the NAS community, which has focused more on RGB images. \textit{iii) \textbf{Novelty}}: We share novel findings to guide the development of better hyperspectral-specific search techniques and architectures.

Our approach begins by generating a substantial pool of candidate Hyperspectral transformers. Subsequently, we assess the performance of various proxies in their capacity to efficiently identify accurate Hyperspectral image transformers. Lastly, we conduct a comprehensive factor analysis to elucidate the factors influencing the performance of both Hyperspectral image transformers and proxy-based transformer search methods.

To sum up, this paper presents three primary contributions:

\vspace{-3mm}

\begin{enumerate}[label=\Roman*.]
\item We introduce HyTAS, the pioneering benchmark for Transformer Architecture Search in HSI classification, comprising $\textbf{2000}$ distinct Hyperspectral Image transformer architectures.

\item We conduct comprehensive benchmarking using $\textbf{12}$ proxies across $\textbf{5}$ datasets. Our benchmarking demonstrates the capability to achieve superior performance over a well-established, human-crafted counterpart within minutes.

\item We analyze various factors influencing the effectiveness of Transformer architecture search. Our findings include: \textit{i)} Proxies demonstrate a bias towards larger models, \textit{ii)} Proxies operate without necessitating real data input, \textit{iii)} Proxies synergistically complement each other in predicting the accuracy of input transformers, suggesting promising avenues for future research.
\end{enumerate}

\section{Related Work}

\subsection{Transformer Architecture Search via Proxies}

Neural Architecture Search (NAS) customizes optimal architectures for specific tasks \cite{baker2016designing, zoph2016neural, liu2018progressive, pham2018efficient}. Transformer Architecture Search (TAS) is gaining popularity alongside transformer-based model advancements \cite{chen2021autoformer, chen2021glit, su2022vitas, dss}. However, these methods are computationally intensive, often taking days or months to converge \cite{li2023zero, white2023neural}. Many researchers lack resources; for instance, this study employs only a single GPU. An alternative approach involves using proxy functions, which swiftly assess architecture fitness without training \cite{li2023zero, snip, synflow, logsynflow, li2023zico}. In this paper, we adapt proxy methods to quickly search over a large database of Hyperspectral transformers.

\subsection{Hyperspectral Image Transformers}

Hyperspectral image classification involves processing input cubes with significantly more channels (\textit{e.g.}, $200$ vs. $3$) and lower spatial resolution than conventional RGB imaging \cite{li2019deep}. With the emergence of Vision-Transformers (ViT) \cite{ViT}, manual design of transformer architectures for Hyperspectral image classification has become common practice. For instance, Hong et al. \cite{hong2021spectralformer} introduce SpectralFormer, adapting ViT for HSI classification, while Sun et al. \cite{sun2022spectral} and Zhou et al. \cite{zhou2023locality} enhance it with spatial-local attention modules. However, manual network design is laborious and resource-intensive. Therefore, in this study, we propose HyTAS to automate Hyperspectral image transformer design.
\section{HyTAS Benchmark}
In practice, the optimal architecture is determined by its performance, often measured by the highest test accuracy for classification tasks. However, in TAS proxies, where training is not involved, a proxy measure is needed to rank architectures sampled from a search space based on their initial model weights or one-step gradients. This proxy score serves as the metric for estimating the performance of each architecture. 

\begin{figure}[t]
    \centering
    \includegraphics[width=1.0\textwidth]{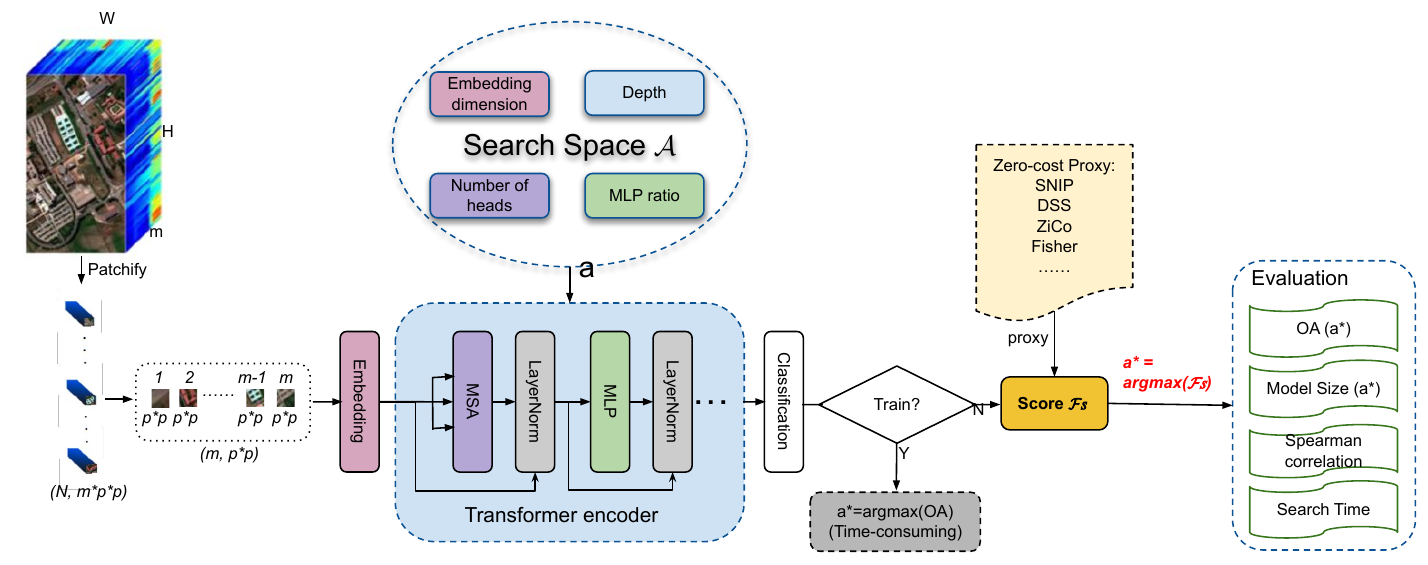}
    \caption{Diagram of TAS proxies for HSI classification: (1) Patchify the image and tokenize each patch along its spectra. (2) Randomly sample architectures from a designed search space. (3) Select a proxy to compute scores for sampled architectures after passing a batch of input. (4) Use the scores to rank architectures and choose the one with the highest score to retrain. Proxy evaluation metrics include test overall accuracy, model size, Spearman correlation between proxy scores and test OA, and search time.}
    \label{fig:diagram}
\end{figure}

The process of TAS proxy for HSI classification is illustrated in Fig.~\ref{fig:diagram}, containing four main steps: 1). Patchify HSI image and tokenize each patch along its spectra, then randomly sample a batch of input. 2). Design a search space and sample architectures from it. 3). Select a proxy to estimate the sampled architectures and rank them by their proxy scores. 4). Evaluate the performance of search proxies. Formally, the objective of TAS proxy is formulated as follows:
\begin{equation}
a^* = \arg\max_{a \in \mathcal{A}} \operatorname{OA}\left(a\right) \equiv a^* = \arg\max_{a \in \mathcal{A}} \mathcal{F}_s\left(a\right)
\label{eq:objective}
\end{equation}
where $a$ is a transformer architecture sampled from the search space $\mathcal{A}$, OA and $\mathcal{F}_s$ denote the test overall accuracy and a scoring function, respectively. The objective aims to discover the optimal architecture $a^*$ through a scoring function instead of training all models. Designing a good search space $\mathcal{A}$ and a suitable proxy scoring function that accurately estimates architectures' performance is crucial for TAS proxies.

\subsection{Benchmark Proxies}
Optimizing Equation \ref{eq:objective} relies on employing an appropriate proxy scoring function $\mathcal{F}_s$. In this section, we provide a concise overview of various existing proxies commonly utilized within the NAS and TAS communities. The mathematical equation of each proxy as well as further details are presented in Section \textcolor{red}{7.1} in our supplementary material.

\textbf{GradNorm} \cite{gradnormfisher}: aggregates the Euclidean norms of each layer's gradients after processing a single batch of training data.

\textbf{SNIP} \cite{snip}: calculates a saliency metric by incorporating both gradients and weights, facilitating parameter pruning in neural networks.

\textbf{GraSP} \cite{grasp}: considers both the first-order and second-order derivatives of the network to preserve training dynamics instead of the loss at the beginning of training.

\textbf{Synflow} \cite{synflow}: a modification of SNIP that eliminates the absolute expression to address layer collapse during parameter pruning.

\textbf{LogSynflow} \cite{logsynflow}: addresses the issue of neglecting the significance of weights due to gradients in Synflow, by scaling down the gradients with a logarithmic function before summing up the contributions of their weights.

\textbf{Fisher}: initially presented by \cite{theis2018faster} to achieve more runtime-efficient neural architectures, and utilized as a pruning metric at initialization by \cite{turner2019blockswap}. More recently, \cite{gradnormfisher} applied this metric to aggregate all gradients of the activations for scoring a network.

\textbf{JacobCov} \cite{jacobian}: leverages gradients over the input data instead of parameters or activations.

\textbf{NASWOT} \cite{naswot}: computes the network score according to its kernel matrix of the binary activation patterns between a batch of training inputs.

\textbf{DSS} \cite{dss}: specifically designed for transformer architectures via combining synaptic diversity scores of MSA and synaptic saliency scores of MLP modules.

\textbf{CRoZe} \cite{croze}: measures the consistency of features, parameters, and gradients between perturbed and clean samples.

\textbf{T-CET} \cite{tcet}: incorporates the orthogonality of gradient-based (SSNR) and activation-based (NASWOT) compressibility to score an architecture.

\textbf{ZiCo} \cite{li2023zico}: considers both absolute mean and standard deviation of gradients, emphasizing their influence on convergence rate and generalization capacity.

\subsection{A New Proxy: ZiCo$^{++}$}
Inspired by ZiCo's principle of favoring networks with high absolute mean and low standard deviation gradient values for faster convergence and improved generalization, we introduce ZiCo$^{++}$, a novel proxy. ZiCo$^{++}$ enhances network expressive capacity by optimizing the mean and minimizing the standard deviation of Fisher information of activations across input samples. Additionally, to address the issue of the final score's heavy reliance on the number of layers in ZiCo, we incorporate a layer decay mechanism during score aggregation in ZiCo$^{++}$. The mathematical formulation is as follows:
\begin{equation}
\scriptsize
\begin{aligned}
& ZiCo^{++} \triangleq \sum_{i=1}^{n-1} ZiCo^{++}_i + \sum_{i=n}^{N-1} \frac{1}{i-n+1}ZiCo^{++}_i + ZiCo^{++}_N  \quad \text{s.t.} \\
& ZiCo_i^{++} \triangleq \log \left(\sum_{\omega \in \theta_l} \frac{\mathbb{E}\left[\left(\nabla_\omega L\left(X_{i}, y_{i} ; z_i\right)* z_i \right)^2\right]}{\sqrt{\operatorname{Var}\left(\left(\nabla_\omega L\left(X_{i}, y_{i} ; z_i\right) * z_i\right)^2\right)}}\right)
\end{aligned}
\end{equation}
 where ZiCo$^{++}_i$ and ZiCo$^{++}_N$ denote the ZiCo$^{++}$ score of $i\_th$ and the last layer, respectively. $n$ is a tunable parameter representing the starting layer for layer decay, such that $n$ is constrained by $n \leq \text{min\_depth}*4+1$, with $\text{min\_depth}$ denoting the minimal depth defined in the search space. In our experiments, we set $\text{min\_depth}=4$ and $n=6$ to preserve the ZiCo$^{++}$ score for the first embedding and the last projection layer, as well as for the first block containing two MSA layers and two MLP layers.

\subsection{Search Space Design} 
A transformer architecture contains an encoder and decoder, but for our classification task, we focus on encoder-only transformers. Typically, a transformer encoder consists of a series of blocks, each containing a multi-head self-attention (MSA) and a multi-layer perceptron (MLP) module. Therefore, the search space should account for four key elements: embedding dimension ($embed\_dim$), number of blocks ($depth$), number of attention heads ($num\_heads$), and MLP embedding ratio ($mlp\_ratio$) for each block. These components significantly impact model performance. 


\begin{table}[]
\centering
\setlength{\tabcolsep}{5pt} 
\renewcommand{\arraystretch}{1.2} 
\begin{tabular}{@{}lcccc@{}}
\toprule
\textbf{Components}       & $\textbf{\textit{depth}}$ & $\textbf{\textit{embed\_dim}}$ &$\textbf{\textit{num\_heads}}$ &$\textbf{\textit{mlp\_ratio}}$    \\
\midrule
\textbf{Choices}       & (4, 10, 1)     & (32, 240, 16)      & (3, 6, 1)  & (1, 6, 1)         \\ 
\bottomrule
\end{tabular}
\caption{The search space across all datasets. It includes $depth$, $embed\_dim$, $num\_heads$, and $mlp\_ratio$, with values indicating initial, final, and intervals. Different numbers of heads and MLP ratios are utilized for each block. Notably, the total number of layers equals $depth * 4 + 2$.}
\label{tab:search_space}
\end{table}

\vspace{-4mm}

Drawing inspiration from the effectiveness of manually crafted models proposed by \cite{hong2021spectralformer, zhou2023locality}, we define our search space in Table \ref{tab:search_space}. We randomly sample 2000 subnetworks from the search space for our experiments.

\section{Experimental Setup}
\paragraph{Datasets} We conduct experiments on five well-established HSI datasets:
\begin{enumerate}
    \item[\textit{i.}] Indian Pines~\cite{hyperdata}: $224$ spectral bands, $145\times145$ spatial resolution, $16$ classes, $695$ training, and $9k$ testing samples.
    \item[\textit{ii.}] Houston2013~\cite{houston2013}: $144$ spectral bands, $349\times1905$ spatial resolution, $15$ classes, $2k$ training, and $12k$ testing samples.
    \item[\textit{iii.}] Pavia University (PaviaU)~\cite{hyperdata}: $103$ spectral bands, $610\times340$ spatial resolution, $9$ classes, $3k$ training, and $40k$ testing samples.
    \item[\textit{iv.}] Kennedy Space Center (KSC)~\cite{hyperdata}: $176$ spectral bands, $518\times620$ spatial resolution, $13$ classes, $195$ training, and $5k$ testing samples.
    \item[\textit{v.}] Salinas scene~\cite{hyperdata}: $224$ spectral bands, $512\times217$ spatial resolution, $16$ classes, $800$ training, and $53k$ testing samples.
\end{enumerate}
Indian Pines and Salinas contain mostly agricultural vegetation, and the remaining three datasets consist of non-agricultural vegetation, such as roads and buildings.
More details are presented in Section \textcolor{red}{7.2} in our supplementary material. For data-dependent search proxies, we randomly sample a batch of training data with a batch size of 64, consistent with the batch size used for retraining.
\paragraph{Metrics} We evaluate architecture performance with test Overall Accuracy (OA), indicating correctly predicted samples over all samples. Spearman correlation ($\rho$) measures the correlation between proxy scores and final test OA after training all sampled networks. We also compare Search Time (ST) and proposed Model Size (MS) for each proxy, calculated using the following equation from~\cite{anonymous2023seeking}:
\[
\scalebox{0.75}{$
MS(a, b, c, d)=1,539 d+a\left(4 d+256 c d+192 c+5 d+7,680+2 b d^2+b d+d\right)+2 d+\# \text { class } \times(d+1)
$}
\]
Here, $a$, $b$, $c$, and $d$ represent the depth, MLP ratio, number of heads, and embedding dimension, respectively. $\#\text{class}$ is the number of classes in the dataset.

\clearpage
\section{HyTAS Analysis}
This section presents experiments to answer the following research questions:

\begin{itemize}
\item RQ1: How do proxies perform on HSI classification? 
\item RQ2: What factors influence Hyperspectral image transformer performance? 
\item RQ3: What factors influence the proxy scores? 
\item RQ4: Are proxies complementary? 

\end{itemize}

\subsection{RQ1: How Do Proxies Perform on HSI Classification?} 

Previous studies have shown the varying performance of proxies under different search space constraints \cite{white2023neural, li2023zero}. To address RQ1, we conduct experiments under full and constrained search space conditions.

\paragraph{\textbf{Full Search Space}} 
We compare proxies across all (2000) subnetworks sampled from the entire search space on five datasets. We include two reference results: SpectralFormer as the human-crafted transformer, and Oracle as the upper bound. Results are presented in Table~\ref{tab:results_1}. We make four observations:

\begin{table}[h]
\centering
\scriptsize
\renewcommand{\arraystretch}{1.2} 
\begin{tabular}{@{}lccccccccccccccc@{}}
\toprule
\textbf{Dataset} & \multicolumn{3}{c}{\textbf{Indian Pines}} & \multicolumn{3}{c}{\textbf{Houston2013}} & \multicolumn{3}{c}{\textbf{PaviaU}} & \multicolumn{3}{c}{\textbf{KSC}}  & \multicolumn{3}{c}{\textbf{Salinas}} \\ 
\cmidrule(lr){2-4} \cmidrule(lr){5-7} \cmidrule(lr){8-10} \cmidrule(lr){11-13} \cmidrule(lr){14-16}
\textbf{Proxy} & OA & MS & $\rho$ &  OA & MS & $\rho$ & OA & MS & $\rho$ & OA & MS & $\rho$ & OA & MS & $\rho$   \\ 
\midrule

\textbf{\textit{\#Flops}}                       & 0.79                        & 26.3 & 0.69          & 0.86                        & 26.3          & 0.66          & 0.89                        & 26.3 & 0.71          & 0.89                        & 26.3 & 0.82          & 0.91                        & 26.3 & \textbf{0.94} \\
\textbf{\textit{GradNorm}}                    & 0.81                        & 26.8 & 0.71          & 0.86                        & 29.8          & 0.70          & \textbf{0.90}               & 29.8 & 0.62          & \textbf{0.90}               & 29.2 & \textbf{0.84} & 0.92                        & 29.8 & 0.87          \\
\textbf{\textit{SNIP}}                          & 0.81                        & 26.8 & 0.70          & 0.87                        & 29.9          & 0.66          & 0.89                        & 26.3 & 0.69          & 0.89                        & 28.1          & 0.83          & 0.93                        & 26.8          & \textbf{0.94} \\
\textbf{\textit{GraSP}}                         & 0.42                        & 0.62 & -0.67         & 0.78                        & 0.94 & -0.63         & 0.78                        & 0.62 & -0.67         & 0.47                        & 0.62 & -0.79         & 0.39                        & 1.23 & -0.88         \\
\textbf{\textit{Synflow}}                       & 0.79                        & 7.35 & 0.60          & 0.84                        & 7.35 & 0.53          & \textbf{0.90}               & 7.35 & \textbf{0.73} & 0.87                        & 7.35 & 0.71          & 0.92                        & 7.35 &
 0.91          \\
\textbf{\textit{LogSynflow}}                    & 0.79                        & 7.35 & 0.60          & 0.84                        & 7.35          & 0.52          & \textbf{0.90}               & 7.35 & \textbf{0.73} & 0.87                        & 7.35          & 0.68          & 0.92                        & 7.35          & 0.89          \\
\textbf{\textit{Fisher}}                        & 0.81                        & 16.3 & 0.61          & 0.87                        & 18.5          & 0.64          & \textbf{0.90}               & 29.3 & 0.61          & 0.89                        & 28.1          & 0.79          & 0.92                        & 9.83          & 0.92          \\
\textbf{\textit{JacobCov}}                    & \textbf{0.82}               & 11.9 & 0.58          & \textbf{0.88}               & 21.4          & 0.72          & 0.84                        & 5.63 & 0.40          & \textbf{0.90}               & 29.8          & 0.70          & 0.90                        & 13.7          & 0.52          \\
\textbf{\textit{NASWOT}}                        & 0.79                        & 26.3 & 0.69          & 0.87                        & 23.1          & 0.71          & 0.89                        & 23.1 & 0.62          & 0.89                        & 26.3          & 0.83          & 0.91                        & 26.3          & 0.87          \\
\textbf{\textit{T-CET}}                   & 0.79                        & 26.3     & 0.63          & 0.86                        &     26.3          & 0.71          & 0.89                        &   26.3   & 0.38          & 0.89                        &     26.3          & 0.74          & \textbf{0.94}               &  29.9             & 0.65          \\
\textbf{\textit{CRoZe}}                         & 0.79                        & 26.3 & 0.46          & 0.87                        & 26.7          & 0.54          & 0.84                        & 26.4 & 0.18          & 0.89                        & 29.3          & 0.55          & 0.91                        & 28.1          & 0.42          \\
\textbf{\textit{DSS}}                           & 0.80                        & 25.9 & 0.60          & 0.84                        & 7.35          & 0.53          & 0.89                        & 9.89 & \textbf{0.73} & 0.89                        & 10.6          & 0.72          & 0.92                        & 29.8          & 0.91          \\
\textbf{\textit{ZiCo}}                          & 0.81                        & 26.8 & 0.52          & 0.86                        & 26.3          & 0.62          & 0.84                        & 25.6 & 0.26          & 0.89                        & 26.3          & 0.63          & 0.91                        & 26.3          & 0.50          \\ 
\textbf{\textit{ZiCo$^{++}$}}                      & 0.81                        & 25.6     & \textbf{0.73} & 0.86                        & 29.3              & \textbf{0.75} & 0.84                        & 25.6     & 0.62          & 0.89                        &  29.3             & 0.83          & \textbf{0.94}               &  29.3             & 0.86          \\ \hdashline
\textbf{\textit{SpectralFormer}}                & 0.79                        &  0.48    & -           & 0.85                        &  0.48             & -           & 0.85                        &   0.48   & -           & 0.86                        &  0.48             & -           & 0.89                        &    0.48           & -           \\ 
 \textbf{\textit{{\color[HTML]{FE0000} Oracle}}} &  0.85&  8.42    & -           & 0.89 &    9.23           & -           & 0.91&   3.20   & -           & 0.91 &   8.43            & -           & 0.96 &   20.5            & - \\

\bottomrule
\end{tabular}
\caption{Comparison of search results among proxies, presenting test Overall Accuracy (OA) and corresponding model size (MS) determined by each proxy, as well as their Spearman correlation ($\rho$) between proxy scores and final test OA after training. The evaluation is conducted over a search space of 2000 subnets across five HSI datasets. MS values are scaled by $10^6$. T-CET is computed with SNIP and NASWOT\cite{tcet}.}
\label{tab:results_1}
\end{table}

\begin{enumerate}[label=\textbullet]
  
  \item The effectiveness of proxies varies across datasets. JacobCov achieves the highest OA for Indian Pines, Houston2013, and KSC datasets, while GradNorm and T-CET demonstrate the highest OA of $0.90$ and $0.94$ for the PaviaU and Salinas datasets, respectively.

  \item Some proxies may yield high OA but exhibit low Spearman correlation. For instance, JacobCov achieves the highest OA for three datasets but with considerably lower Spearman correlations compared to the best ones, emphasizing the importance of considering both OA and Spearman correlation for comprehensive assessment.

  \item ZiCo$^{++}$ does not outperform ZiCo in terms of OA across all datasets except for Salinas. However, it significantly exceeds ZiCo in Spearman correlation across all datasets, achieving the highest Spearman correlation among all proxies for the Indian Pines and Houston datasets.

  \item Apart from OA, model size, and $\rho$, searching efficiency is crucial for evaluating proxies. Table \ref{tab:search_time} compares search times of all proxies. Except for ZiCo and CRoZe, the search times for the other proxies are remarkably similar. ZiCo's extended search time can be attributed to the computation of scores across individual samples, requiring gradient calculations for each sample rather than utilizing batch mean gradients.

\end{enumerate}

\begin{table}[htbp]
\centering
\tiny
\setlength{\tabcolsep}{2.5pt} 
\renewcommand{\arraystretch}{1.2} 
\begin{tabular}{@{}lcccccccccc@{}}
\toprule
\textbf{Proxy}       & \textbf{\textit{GradNorm}} & \textbf{\textit{SNIP}} & \textbf{\textit{Synflow}} & \textbf{\textit{Fisher}} & \textbf{\textit{JacobCov}} & \textbf{\textit{NASWOT}} & \textbf{\textit{CRoZe}} & \textbf{\textit{DSS}} & \textbf{\textit{ZiCo}} & \textbf{\textit{ZiCo$^{++}$}} \\
\midrule
\textbf{Search time (h)} & 0.12 & 0.12 & 0.12 & 0.20 & 0.12 & 0.14 & 0.38 & 0.17 & 0.84 & 0.14 \\
\bottomrule
\end{tabular}
\caption{Comparison of search times for each proxy across a search space of 2000 subnets on the Indian Pines dataset.}
\label{tab:search_time}
\end{table}

\vspace{-4mm}

\paragraph{\textbf{Constrained Search Space}} 
We visualize Spearman correlations and proposed OA across varying model size constraints in Fig.~\ref{fig:model_sizes}. The depicted range represents the minimum and maximum model sizes within the search space. For instance, the range from 0 to 0.5e7 signifies model sizes from 0 to 5M, and from 0.5e7 to 1e7, it represents sizes from 5M to 10M. We make three observations: 

\begin{figure}[h]
\centering
\includegraphics[width=1.0\textwidth]{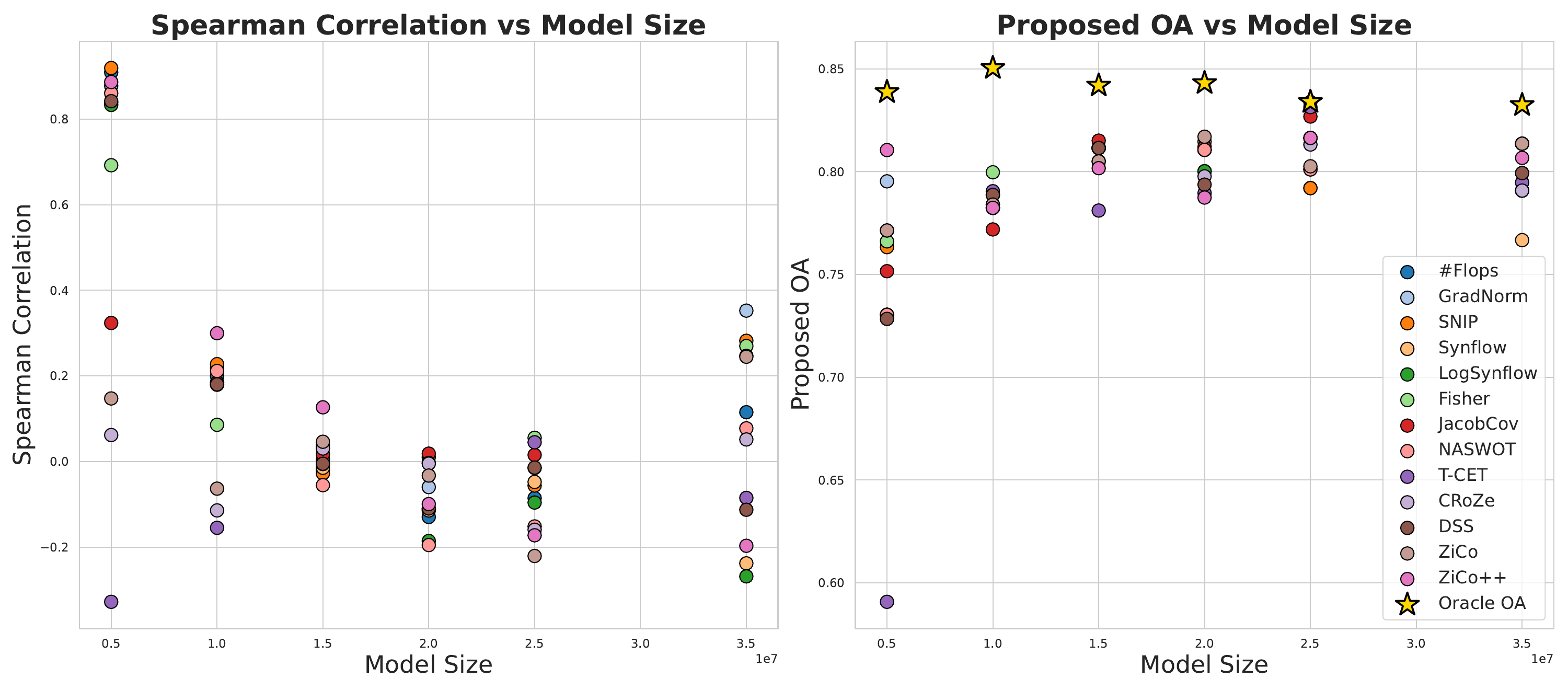}
\caption{Spearman correlation and proxy-proposed OA across different model size constraints on the Indian Pines dataset.}
\label{fig:model_sizes}
\end{figure}

\begin{enumerate}[label=\textbullet]

    \item The Spearman correlation between proxies and OA decreases with increasing model size, indicating higher efficacy of proxies for smaller models.
    \item The Oracle OA increases for models smaller than 5M and remains relatively stable thereafter, with the optimal OA observed at 5M.
    \item Most proxies favor complex models and struggle when simpler models are more effective, suggesting their limited capability to filter out underperforming models.
\end{enumerate}

\textbf{Takeaway $\protect\circled{1}$:} Most proxies can identify a transformer with better accuracy than the human-crafted counterpart, SpectralFormer. However, the proposed architectures are more complex than necessary, when compared to the oracle.

\subsection{RQ2: What Factors Influence Hyperspectral Image Transformer Performance? }

In RQ1, we show that proxies identify a highly accurate, yet complex model. However, is this an optimal Hyperspectral transformer? To answer this, in this section, we analyze the relationship between the performance and the architectural factors. We have three key observations: 

\begin{figure}[h]
    \centering
    \includegraphics[width=0.9\textwidth]{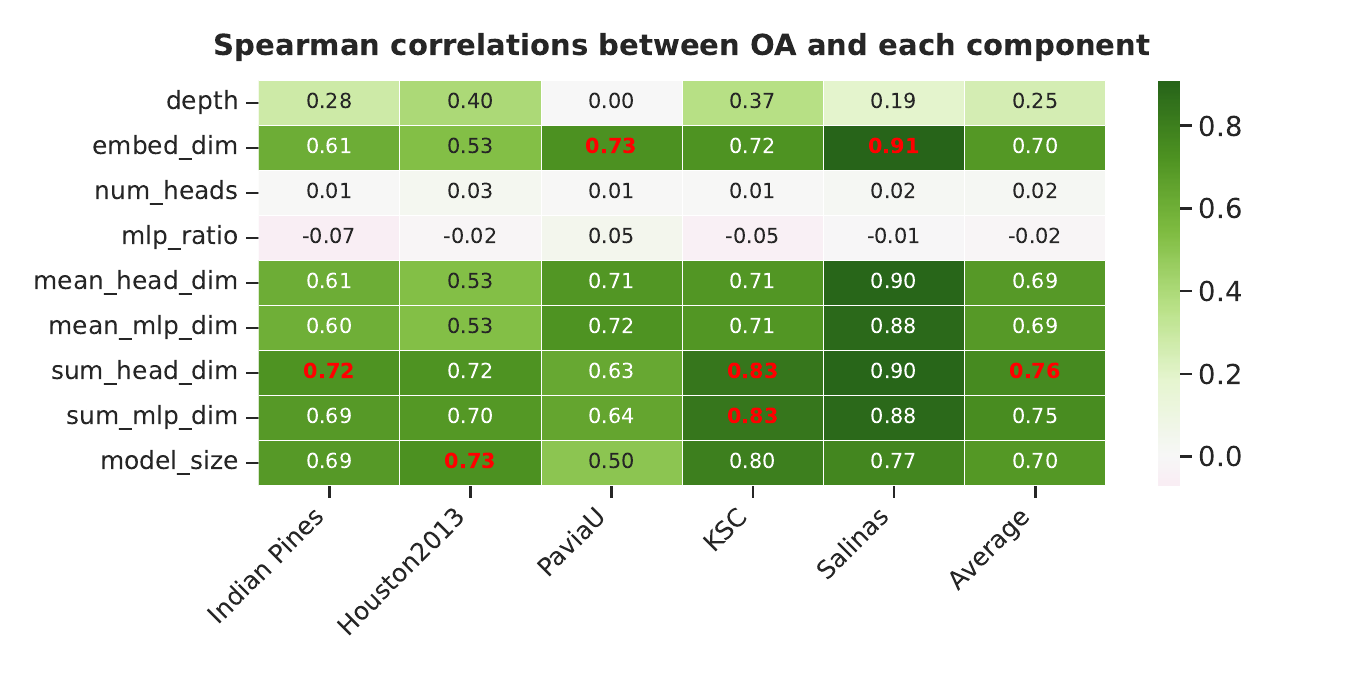}
    \caption{Spearman correlations between final test OA and components of the search space across five datasets. $num\_heads$ and $mlp\_ratio$ exhibit minimal correlation, while $embed\_dim$, $sum\_head\_dim$, and $sum\_mlp\_dim$ show high correlation across all datasets.}
    \label{fig:factors_oa_hm}
\end{figure}

\begin{figure}
    \centering
    \includegraphics[width=1.0\textwidth]{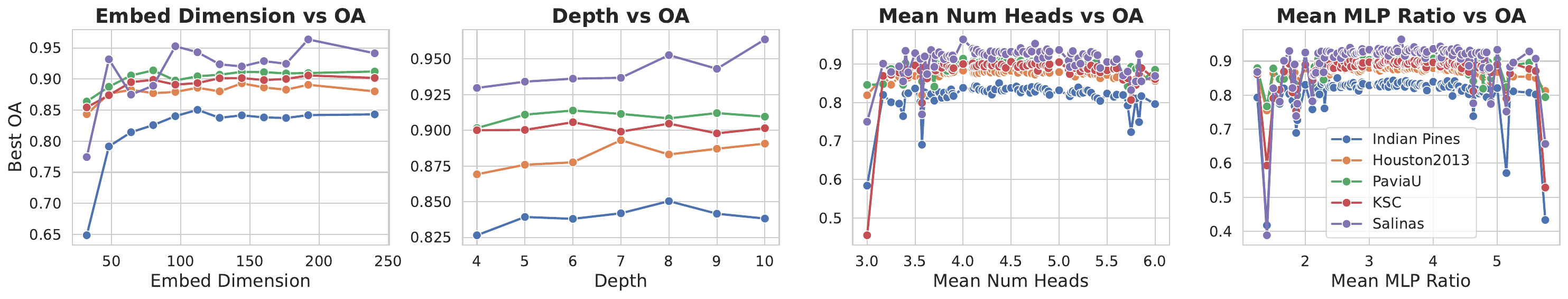}
    \caption{The best OA corresponding to different values of diverse factors, including the embedding dimension ($embed\_dim$), the depth ($depth$), the average number of heads ($mean\_heads\_num$), and the average MLP ratio ($mean\_mlp\_ratio$).}
    \label{fig:factor_oa}
\end{figure}

\begin{enumerate}[label=\textbullet]

    \item Fig.~\ref{fig:factors_oa_hm} presents the Spearman correlation between OA and: depth ($depth$), embedding dimension ($embed\_dim$), number of heads ($num\_heads$), and MLP ratio ($mlp\_ratio$). Notably, factors such as the number of heads and MLP ratio show minimal correlation with overall model performance across all datasets. Conversely, the influence of depth and embedding dimension varies across datasets, with embedding dimension exhibiting particularly high correlations across all datasets, reaching 0.91 on Salinas dataset.
    \item Fig. \ref{fig:factors_oa_hm} presents average head embedding dimension ($mean\_head\_dim$), average MLP dimension ($mean\_mlp\_dim$), sum of all head dimensions 
    
    ($sum\_head\_dim$), sum of all MLP dimensions ($sum\_mlp\_dim$), and model size. Notably, average head dimension and average MLP ratio exhibit OA impacts similar to the embedding dimension, while total head dimension and total MLP ratio aggregate influence from both embedding dimension and depth. However, for datasets like PaviaU, where OA relies solely on embedding dimension, the impact of total head dimension diminishes. Conversely, for datasets like Indian Pines, Houston2013, and KSC, where OA correlates with both embedding dimension and depth, Spearman correlation of total head dimension increases. This suggests that if proxy scores correlate solely with one factor, they may not be robust indicators.
    \item In Fig. \ref{fig:factor_oa}, we visualize the highest OA of different values of each component across all datasets. Optimal embedding dimension and depth vary across datasets. As an example, in Indian Pines, OA increases with the embedding dimension, peaking at $embed\_dim=128$ and remaining stable thereafter. Similarly, $depth=8$ results in the highest OA, with no further improvements observed when increasing the depth to 9 or 10. Additionally, the impact of the average number of heads and the average MLP ratio on model performance is showcased in the right two subplots, where mean number of heads of $4-5$ and mean MLP ratio of $3-4$ yield relatively stable and high OA.
\end{enumerate}

\textbf{Takeaway $\protect\circled{2}$:} Across datasets, embedding dimension and depth exhibit consistent sensitivity, though with slight variations. Furthermore, while larger models generally achieve better performance, optimal networks do not necessarily have to be the largest or most complex ones.

\subsection{RQ3: What Factors Influence the Proxy Scores?} 
To answer this question, we analyze the sensitivity of proxy scores to architecture components, input data, and module types separately.

\paragraph{\textbf{Sensitivity to Architectural Components}} 

\begin{figure} 
    \centering
    \includegraphics[width=1.1\textwidth]{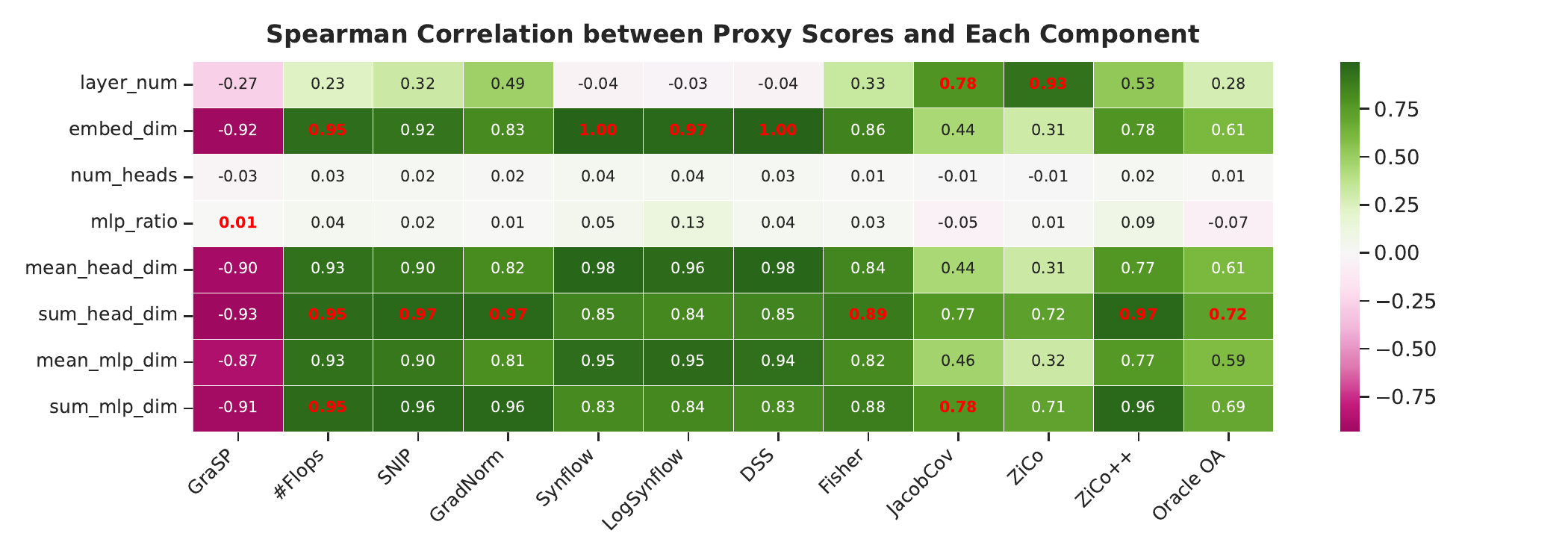}
    \caption{Spearman correlations between proxy scores and architecture components on Indian Pines dataset.}
    \label{fig:factors_inds}
\end{figure}

We examine Spearman correlations between proxy scores and each component on Indian Pines dataset in Fig. \ref{fig:factors_inds}. Additionally, Spearman correlations between OA and each component are shown in the last column as a comparison. Depth and embedding dimension significantly impact scores of most proxies. Specifically, $\#$Flops, SNIP, and GradNorm display a strong correlation (over 0.95) with the sum of heads' dimensions, surpassing the correlation with OA. Furthermore, Synflow, LogSynflow, and DSS show perfect correlation with embedding dimension and minimal correlation with depth due to taking sign of model parameters, which also exceeds the correlation with OA. Interestingly, removing sign operations improves the performance of Synflow, LogSynflow, and DSS. Details are shown in Section \textcolor{red}{7.3} in our supplementary material. These discrepancies in the correlation between OA and proxy scores result in a performance gap between proxies and oracle results.

\vspace{-8mm}

\paragraph{\textbf{Sensitivity to Input Data}} 

Considering Synflow, LogSynflow, NASWOT, and DSS are data-agnostic, we explore the sensitivity of remaining proxies to input data. To investigate this, we conduct search experiments using one batch of random input data across the remaining proxies. The results, presented in Table \ref{tab:random}, reveal that performing the search with random inputs yields outcomes very similar to those obtained with the Indian Pines dataset. This suggests that the majority of the proxies exhibit negligible dependency on specific input data.

\begin{table}[]
\centering
\scriptsize
\setlength{\tabcolsep}{7.8pt} 
\renewcommand{\arraystretch}{1.2} 
\begin{tabular}{@{}lcccc@{}}
\toprule

\multicolumn{1}{l}{\multirow{2}{*}{\textbf{Proxy}}}      & \multicolumn{2}{c}{\textbf{OA}} & \multicolumn{2}{c}{\textbf{$\rho$}} \\
\cmidrule(lr){2-3} \cmidrule(lr){4-5}
         & \textbf{Indian Pines}             & \textbf{Random Input}             &  \textbf{Indian Pines}             & \textbf{Random Input}          \\
\midrule
\textbf{\textit{GradNorm}}     & 0.81  &  0.81          & 0.71                         &  0.71              \\
\textbf{\textit{SNIP}}           & 0.81    &  0.81                & 0.70                 & 0.69 (-0.01)                \\
\textbf{\textit{GraSP}}          & 0.42     & 0.44 (+0.02)        & -0.67                        &   -0.70 (-0.03)             \\
\textbf{\textit{Fisher}}         & 0.81     &  0.82 (+0.01)       & 0.61                        &  0.55 (-0.06)              \\
\textbf{\textit{JacobCov}}     & 0.82      &  0.79 (-0.03)         & 0.58                    &  0.68 (+0.1)            \\
\textbf{\textit{ZiCo}}           & 0.81    &  0.79 (-0.02)          & 0.52                      &    0.52            \\
\bottomrule
\end{tabular}
\caption{Comparison between searching with a batch of Indian Pines inputs and random inputs.}
\label{tab:random}
\end{table}

\begin{figure}
    \centering
    \includegraphics[width=1.0\textwidth]{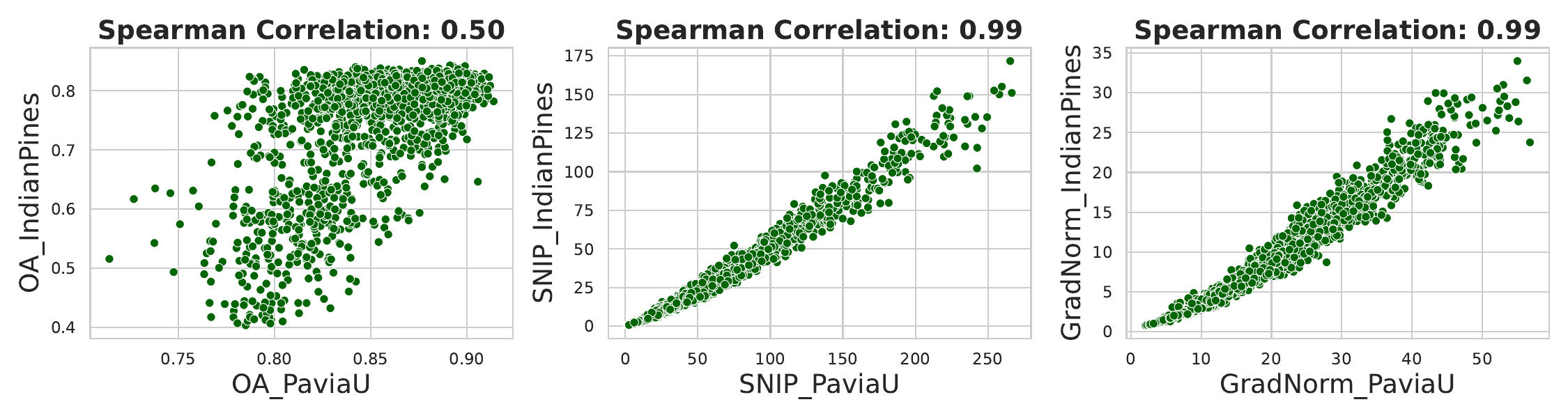}
    \caption{Comparisons of OA, SNIP, GradNorm of same subnetworks between PaviaU and Indian Pines datasets.}
    \label{fig:pavia_indian}
\end{figure}

Furthermore, we compare OA, SNIP, and GradNorm scores from the same models between PaviaU and Indian Pines datasets in Fig. \ref{fig:pavia_indian}. The plots of SNIP and GradNorm indicate a perfect correlation between scores across different datasets, suggesting that the proxies are largely independent of the input data. However, the plot of OA shows significant variation between the two datasets, indicating that the same architecture yields very different performances depending on the dataset. This finding further emphasizes the gap between proxies and model performance.

\vspace{-7mm}

\paragraph{\textbf{Sensitivity to Module Type (MSA \textit{vs.} MLP)}} 

Given the distinct structures of MSA and MLP modules within a transformer architecture, we hypothesize that the effectiveness of the two modules for proxy scores might differ significantly. To validate our assumption, we separately compute DSS, Synflow, and SNIP scores for MSA and MLP modules. Our analysis reveals a significant score scale difference between MSA and MLP. For instance, for DSS scores, MSA ($2472 \pm 259$) is notably higher than MLP ($268 \pm 366$), indicating the predominant contribution of MSA modules to final proxy scores, with minimal influence from MLP modules. To address this imbalance, we apply a logarithmic transformation to both module scores. The transformed results demonstrate comparable scores for MSA and MLP modules across all three proxies. Detailed scores are presented in Section \textcolor{red}{7.4} in our supplementary material.

Furthermore, we assess the effectiveness of scores with and without logarithmic transformation, as shown in Table \ref{tab:logarithm}. The origin score is represented as $origin\_score = msa\_score + mlp\_score$, and the new score is represented as $logarithm\_score = \log(msa\_score \times mlp\_score)$, where $msa\_score$ represents the cumulative score of all MSA layers, and $mlp\_score$ denotes the cumulative score of all MLP layers. The results indicate improvements with the logarithmic transformation, particularly for Synflow and DSS proxies.

\begin{table}[]
\centering
\scriptsize
\setlength{\tabcolsep}{2.6pt} 
\renewcommand{\arraystretch}{1.2} 
\begin{tabular}{@{}lcccccccc@{}}
\toprule
\multicolumn{1}{l}{\multirow{2}{*}{\textbf{Dataset}}} & \multicolumn{2}{c}{\textbf{\textit{SNIP}}}                                   & \multicolumn{2}{c}{\textbf{\textit{GradNorm}}}                             & \multicolumn{2}{c}{\textbf{\textit{Synflow}}}                                & \multicolumn{2}{c}{\textbf{\textit{DSS}}}          \\              
\cmidrule(lr){2-3} \cmidrule(lr){4-5} \cmidrule(lr){6-7} \cmidrule(lr){8-9}
\multicolumn{1}{c}{}                         & \multicolumn{1}{c}{\textbf{origin}} & \multicolumn{1}{c}{\textbf{logarithm}} & \multicolumn{1}{c}{\textbf{origin}} & \multicolumn{1}{c}{\textbf{logarithm}} & \multicolumn{1}{c}{\textbf{origin}} & \multicolumn{1}{c}{\textbf{logarithm}} & \multicolumn{1}{c}{\textbf{origin}} & \multicolumn{1}{c}{\textbf{logarithm}} \\ 
\midrule
\textbf{Indian Pines}                                 & 0.70              & 0.70                 & 0.71              & 0.71                & 0.69              & \textbf{0.72}                 & 0.69              & \textbf{0.70}                 \\
\textbf{Houston2013}                                  & \textbf{0.66}              & 0.65                          &     0.70                       &     0.70                          & 0.68                       & \textbf{0.73}                 & 0.67                       & \textbf{0.69}                 \\
\textbf{PaviaU}                                       & 0.69                       & \textbf{0.70}                 &     0.62                       &    \textbf{0.64}                           & \textbf{0.69}              & 0.63                          & \textbf{0.70}              & 0.69                          \\
\textbf{KSC}                                          & \textbf{0.83}              & 0.82                 & 0.84                  & 0.84                     & 0.82             & \textbf{0.86}                 & 0.82          & \textbf{0.84}                 \\
\textbf{Salinas}                                      & 0.94                       & 0.94                          & 0.87                       & \textbf{0.90}                 & \textbf{0.93}              & 0.90                          & 0.94                       & 0.94       \\ 
\bottomrule                 
\end{tabular}
\caption{Comparison of Spearman correlations between proxy scores and OA, with and without logarithmic transformation.}
\label{tab:logarithm}
\end{table}

\vspace{-5mm}
\textbf{Takeaway $\protect\circled{3}$:} Proxies exhibit a higher correlation with embedding dimension and depth than actual model performance, suggesting limited sensitivity to input data. Enhancing proxies' robustness may involve integrating more input information and decreasing sensitivity to individual architecture components. Furthermore, assessing scores independently for MSA and MLP modules, or employing distinct metrics for each, could improve proxy performance.

\subsection{RQ4: Are Proxies Complementary? }

\paragraph{\textbf{Motivation}.} Our results indicate that TAS proxies are fast but may lack reliability, with search performance varying across datasets. Therefore, TAS proxies solely remain problematic. Some studies have employed proxies for pre-filtering or combined them with other search methods, such as one-shot search, evolutionary algorithms, or Bayesian optimization \cite{xiang2023zero, white2023neural}. Here, we propose an alternative approach to leverage proxies for enhancing search performance.

\paragraph{\textbf{Setup}.} We seek to forecast a network's performance using its architectural structure and proxy scores. By retraining all sampled networks and collecting their proxy scores, we explore the feasibility of predicting a model's OA, alongside exploring the number of training samples needed for accurate predictions. We apply a default Random Forest model, with inputs comprising 2000 samples. Each sample represents a subnetwork sampled from the search space, encompassing attributes such as $depth$, $embed\_dim$, $num\_heads$, $mlp\_ratio$, and various proxy scores including SNIP, GradNorm, Synflow, DSS, ZiCo, and Fisher scores. The target variable is OA, derived from the Indian Pines dataset. Detailed settings are presented in Section \textcolor{red}{7.5} in our supplementary material.

\begin{figure}[htbp]
    \centering
    \begin{subfigure}[b]{0.46\textwidth}
        \includegraphics[width=\textwidth]{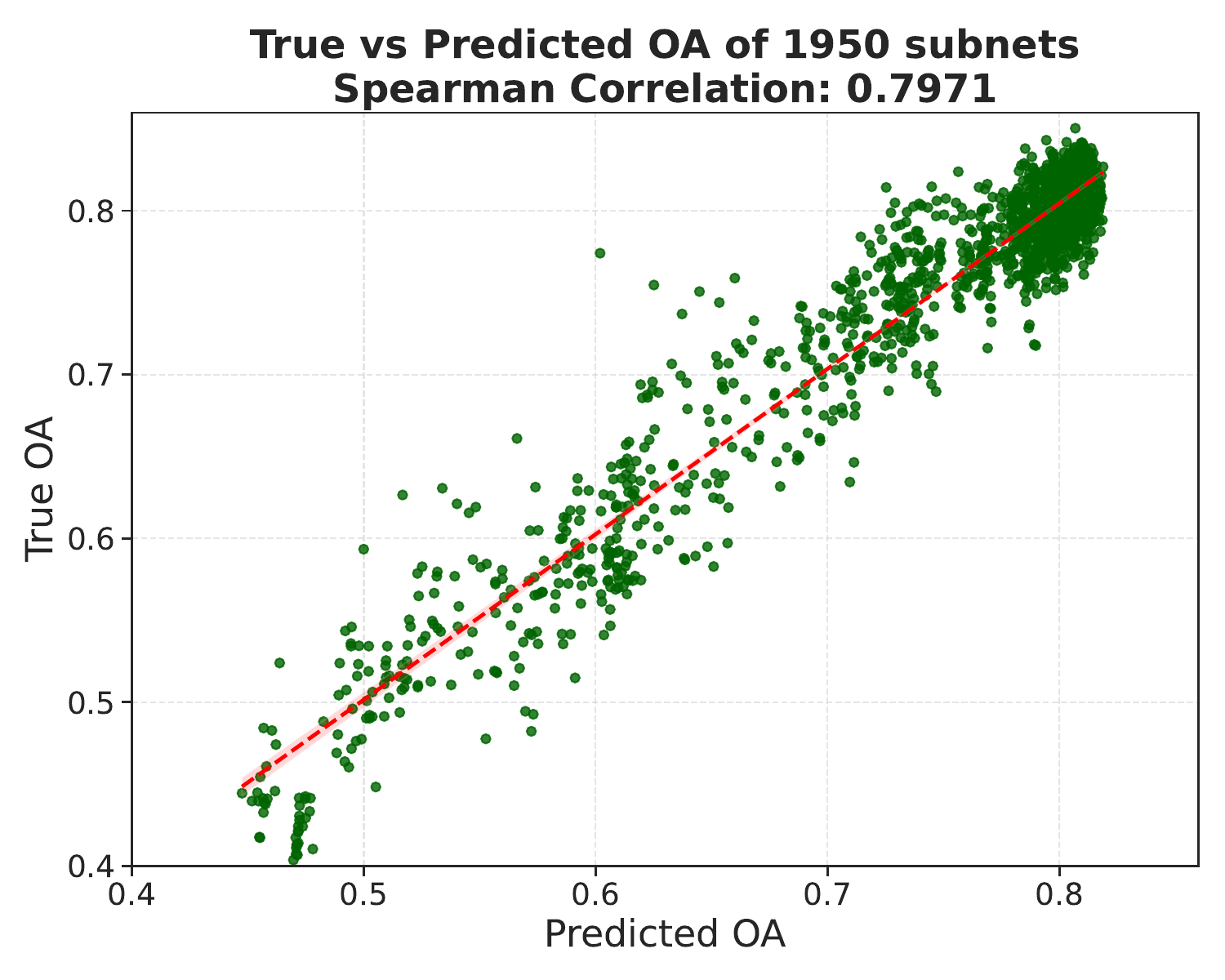}
    \end{subfigure}
    \hfill
    \begin{subfigure}[b]{0.46\textwidth}
        \includegraphics[width=\textwidth]{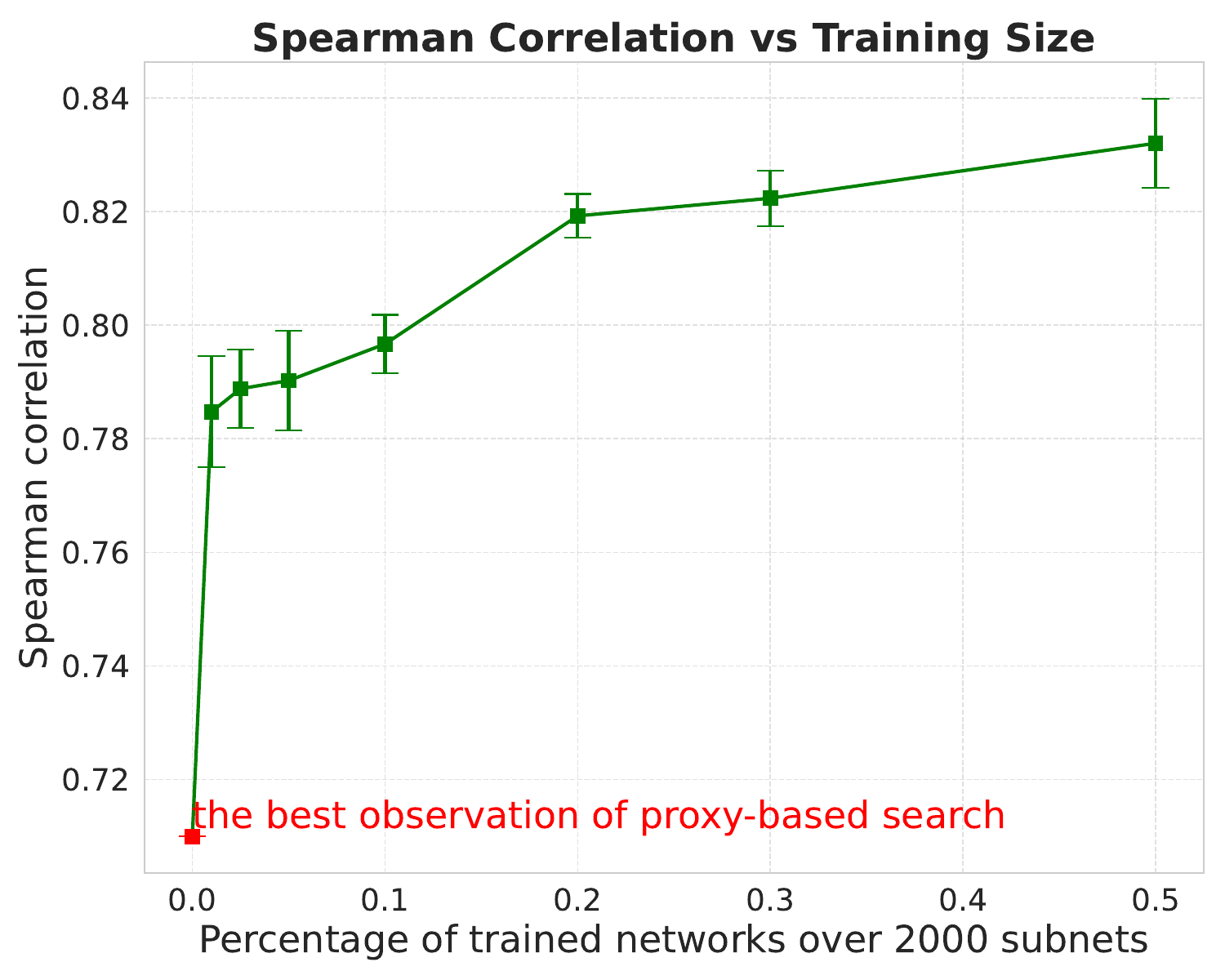}
    \end{subfigure}
    \caption{Predicting subnetworks' performance using Random Forest model.}
    \label{fig:randomforest}
\end{figure}

\paragraph{\textbf{Results}.} Fig. \ref{fig:randomforest} displays the test outcomes. The left subplot presents the predicted OA alongside the actual OA for 1950 networks trained on 50 networks. The Spearman correlation between actual and predicted OA reaches 0.80, surpassing the best correlation obtained from proxies by $9\%$. Additionally, the right subplot demonstrates a correlation improvement with an increase in the number of training networks. Each result represents the mean and standard deviation across five runs with random seeds. Notably, the correlation rises to 0.78, a $6\%$ increase over the untrained scenario, with only 20 networks trained. It is worth noting that training the Random Forest model takes less than one minute with a training size of 0.5x, making the cost almost negligible compared to training the original subnetworks with actual input data.

\textbf{Takeaway $\protect\circled{4}$:} Proxies, while not entirely reliable on their own, can complement other architecture search methods to improve search efficiency or serve as training data for models predicting network performance to improve search accuracy.
\section{Conclusion}

In this study, we introduced HyTAS: a benchmark for Hyperspectral Image Transformer Architecture Search. We defined a search space comprising 2000 Hyperspectral image transformers and evaluated 12 proxies across five HSI datasets. Our key observations are as follows: Most proxies can identify a transformer surpassing the human-crafted architecture, SpectralFormer, within 10 minutes. However, there's a significant performance gap between proxies and optimal results, driven by preferences for larger and more complex models, minimal dependence on input data, and disparities between MSA and MLP modules. Our proposed proxy, ZiCo$^{++}$, demonstrates superior performance compared to ZiCo and other proxies. Furthermore, proxies can enhance the prediction of subnetworks' performance when used as input data. These findings encourage the TAS community to explore new methods for improving search quality and efficiency.

\section*{Acknowledgements}
This work was funded by the Dutch Science Foundation (NWO) under grant 482.20.700. Furthermore, we deeply appreciate the invaluable and constructive feedback offered by our anonymous reviewers and the meta-reviewer. Their insights have greatly enriched our manuscript.


%
%
\bibliographystyle{splncs04}
\bibliography{main}
\end{document}